# Automating Whole Brain Histology to MRI Registration: Implementation of a Computational Pipeline

Maryana Alegro[1,2], Eduardo J. L. Alho[3,4], Maria da Graça Morais Martin[3], Lea Teneholz Grinberg[3,5,6], Helmut Heinsen[5,7], Roseli de Deus Lopes[1], Edson Amaro-Jr[2, 3*], Lilla Zöllei[8,*]

[1]Integrated Systems Laboratory, Escola Politécnica, Universidade de São Paulo, Brazil
[2]Brain Institute, IIEP Albert Einstein, Brazil
[3]Departament of Radiology, Faculdade de Medicina, Universidade de São Paulo, Brazil
[4]Department of Neurosurgery, Faculdade de Medicina, Universidade de São Paulo, Brazil
[5]Departament of Pathology, Faculdade de Medicina, Universidade de São Paulo, Brazil
[6]Memory and Aging Center, Department of Neurology, University of California San Francisco, USA
[7]Morphological Brain Research Unit, Department of Psychiatry, University of Würzburg, Germany
[8]Athinoula A. Martinos Center for Biomedical Imaging, MGH/HST, USA
*Authors contributed equally to the study

**Abstract.** Although the latest advances in MRI technology have allowed the acquisition of higher resolution images, reliable delineation of cytoarchitectural or subcortical nuclei boundaries is not possible. As a result, histological images are still required to identify the exact limits of neuroanatomical structures. However, histological processing is associated with tissue distortion and fixation artifacts, which prevent a direct comparison between the two modalities. Our group has previously proposed a histological procedure based on celloidin embedding that reduces the amount of artifacts and yields high quality whole brain histological slices. Celloidin embedded tissue, nevertheless, still bears distortions that must be corrected. We propose a computational pipeline designed to semi-automatically process the celloidin embedded histology and register them to their MRI counterparts. In this paper we report the accuracy of our pipeline in two whole brain volumes from the Brain Bank of the Brazilian Aging Brain Study Group (BBBABSG). Results were assessed by comparison of manual segmentations from two experts in both MRIs and the registered histological volumes. The two whole brain histology/MRI datasets were successfully registered using minimal user interaction. We also point to possible improvements based on recent implementations that could be added to this pipeline, potentially allowing for higher precision and further performance gains.

1. **Introduction**

Medical imaging has revolutionized neuroscience by providing non-invasive tools to examine the brain *in vivo*, and Magnetic Resonance Imaging (MRI) is currently the most prominent brain-imaging technique as it uses non-ionizing energy for acquiring images with much higher spatial resolution than other modalities, such as computerized tomography (CT). The principle of MRI is based on tissue relaxation properties after radiofrequency excitation. It quickly became popular in neuroscience after its introduction in the late 1980s [Balchandani and T. P. Naidich, 2015; Duyn, 2012]. Since then, MRI has benefited from advances in electronics and superconductor technology, and also from improvements in ultra-high field magnets and faster radio frequency and gradient generators [Polders et al., 2011]. Based on these advances, clinical MRI scanners have also evolved tremendously since their inception, resulting in acquisitions at unprecedented resolution and details in a non-invasive manner [Balchandani and T. P. Naidich, 2015]. The initial 4 x 4 x 4mm$^3$ spatial resolution of the early days evolved to a remarkable 0.5 x 0.5 x 0.5mm$^3$ in currently available clinical MRI acquisitions [Duyn, 2012]. Despite all these developments, the maximum resolution of modern MRI is still far behind that of the average 1 x 1 x 1 μm spatial resolution obtained in histology using a conventional brightfield microscope [Dauguet et al., 2007]. As a result, interpretation of MRI images mostly relies on inferences combining imaging results with clinical outcomes, and surgical and biopsy tissue probing. The problems caused by the lack of a "direct" biological correspondence of the MRI signal is illustrated by the high disagreement between pathologists and radiologists [Hipp et al., 2011] and high inter-observer discrepancy among experienced radiologists [Beach et al., 2012].

Histological examination remains the gold-standard for delineating the borders of microscopic brain structures, including cytoarchitectonic fields and subcortical nuclei, and also for diagnosing the majority of brain conditions [Esiri, 2010]. In theory, further improvements on MRI resolution would enable "in vivo histology" making it possible to diagnose important neurological conditions *in vivo*, including neurodegenerative diseases and the nature of white matter hyperintensities. Furthermore, improved MRI resolution may facilitate accurate cytoarchitectonics and inspection of subcortical structures [Annese, 2012; Aoki et al., 1989; Grinberg et al., 2009]. Since the current high-field clinical MRI scanners already stretch the tolerance of magnet fields and thermal safety limits in humans [Ham et al., 1997; Shellock, 2000], introducing MRI acquisitions with even stronger magnetic fields is probably unfeasible in large scale clinical MRI.

Aiming to improve MRI interpretation in neurology practice using current MRI scanners, we developed a pipeline to enable point-to-point correlation between different MRI acquisitions and the corresponding biological parameters detected directly in the tissue [Grinberg et al., 2009]. We hypothesize that by creating a library of MRI signals with corresponding biological signatures, directly sampling the brain tissue at a microscopic scale will become redundant. Despite being promising, our initial pipeline had two main hinderances:

1) The inherent deformation caused by transforming a 3D volume into 2D brain slices. Brain tissue is soft; procurement and processing inevitably causes local and global deformations. The required sectioning aggravates the problem by generating holes and tearing of tissue on the slides. All these artifacts combined cause vast geometric differences that complicate finding spatial correspondence between histology and MRI data sets. Therefore, we require the use of intricate non-linear deformation models that, in turn, may add more bias. The registration process is further complicated by considerable contrast heterogeneity between MRI and histology, requiring the use of robust similarity measures [Ceritoglu et al., 2010]. Finally, these modalities have different acquisition and sectioning planes, making direct alignment of MRI and histology unfeasible.

2) Generation of large data sets. Human brain sectioning yields hundreds of images and large datasets making the 3D reconstruction very computationally intensive [Amunts et al., 2013]. Most of the steps of our initial pipeline are manual and low throughput.

Here we present a computational pipeline that relies on minimal user interaction and is capable of establishing a direct association between the histology and MRI acquisitions of a full human brain. This pipeline presents a unique combination of known image processing methods to: (a) register all histology images in a data set, (b) recover the original brain geometry, and (c) register the histological volume to the MRI. More importantly, the pipeline is innovative as it includes the original histological colors into the final registration, enabling the use of a broad array of histological methods, such as immunohistochemistry, to detect biological markers of interest. As a proof of concept, we registered two datasets using current off-the-shelf desktop computers and quantified the registration accuracy using the Dice similarity coefficient [Dice, 1945] and the normalized weighted spectral distance (nWSD) [Konukoglu et al., 2012].

**Methods**

## 1.1. MRI acquisition and histology processing
**Data:**

Two *post-mortem* whole human brains from the Brain Bank of the Brazilian Aging Brain Study Group (BBBABSG - [Grinberg et al., 2007]) were used in this study. In this paper we refer to these specimens as Case 01 and Case 02. Each brain was scanned within 6 hours after death (*ex-vivo*), before procurement (*in-situ* – not fixed or removed from the skull) and after (fixed in formalin for 12 weeks) both in the same 3.0 T Philips Achieva scanner using an 8-channel head coil. A 3D isotropic T1-weighted Inversion-recovery 3D Fast Field Echo sequence (TR = 6,3ms; TE = 2,9 ms; TI = 791 ms; 240 x 240; 1x1x1mm voxels; slices oriented to the mid-sagittal plane). The local Ethics Committee approved all procedures and relatives gave their informed consent.

**Tissue processing and histological imaging acquisition:**

Upon procurement, the brains were processed using a celloidin-embedding protocol previously developed by our group [Ewers et al., 2011; Grinberg et al., 2007; Grinberg et al., 2008; Grinberg et al., 2009; Teipel et al., 2008; Teipel et al., 2011; Theofilas et al., 2014]. Briefly, the brains were fixed in 10% buffered formalin for several weeks, dehydrated in graded series of ethanol solutions, soaked in 8% celloidin solution, and celloidin-mounted by means of a vacuum-assisted embedding procedure. The resulting blocks were serially sectioned on a sliding microtome (Polycut, Cambridge Instruments, UK) with a section thickness of 400 µm in the axial (Case 01) and sagittal (Case 02) planes. During the serial sectioning, digital photographs were obtained after each cut with a Canon 5D Mark III camera using a 50mm AutoMakro Olympus lens attached to a copystand. Below, we refer to this set of photographs as "blockface" images. As the distance between the camera and the block remained constant throughout the cutting process, these images remained perfectly aligned with respect to the neighboring slices.

All histological sections were stained using a modified gallocyanin (Nissl) technique optimized for highlighting neuronal bodies [Heinsen and Y. L. Heinsen, 1991; Heinsen et al., 2000] and mounted on slides.

A second set of images, referred to as "histology" below, was acquired by placing each stained slide over a portable light table with uniform white backlight. The slides were photographed using a Fuji Finepix II camera and a 50mm AutoMakro Olympus lens attached to the copystand. All pictures were taken with the same magnification and distance from the camera, thus assuring a standard scale through the whole set. In depth details of the celloidin-embedding and gallocyanin-staining procedures were described in previous publications [Heinsen and Y. L. Heinsen, 1991; Heinsen et al., 2000]. Figure 1 shows some examples of the images used in this study. The first row shows a histology slide, its corresponding blockface image, and a non-aligned MRI slice from Case 01. The second row displays a histology slide, its corresponding blockface image, and a non-aligned MRI slice from Case 02.

### 1.2. Computational pipeline

Whole brain histological images bear several linear and non-linear deformations, intensity inhomogeneity, and other artifacts caused by mechanical and chemical manipulation. Additionally, serial sectioning causes the loss of tridimensional geometry. Stacking slices does not necessarily recover the original 3D shape, yielding a deformed volume. Furthermore, registering such images to their neighboring ones is not effective, as the alignment errors are likely to propagate throughout the volume, yielding a twisted reconstruction [Streicher et al., 1997].

The blockface volume preserves information about the correct slice positioning and is used as a template for correctly aligning the histology. Our strategy is to register each histological image to its corresponding blockface and then to stack them, thus creating a histology volume free of z-effect or banana effect. Z-effect is a registration error observed when histological slices are pair-wise registered and stacked. The registration errors causes the images to rotate around the z-axis yielding a twisted volume [Xu et al., 2015]. Blockface images, however, bear deformations caused by brain removal and fixation and cannot be trusted to recover the brain's original shape.

The brain's original geometry can only be recovered by computing 3D spatial correspondence with the undistorted (*in situ*) MRI. Registration with MRI also has the advantage of overcoming differences in the number of slices and acquisition planes.

We designed our pipeline aiming to perform these tasks with minimum user interaction, and we included all the preprocessing steps necessary for achieving a high-quality registration. The following sections give an in depth description of all the pipeline steps. Figure 2 shows a graphical outline of our proposal. It is split into two main branches, histology processing and MRI processing, which can both run independently. Results obtained after each step are displayed in Figure 3.

The pipeline was developed in Matlab (MathWorks, USA). FreeSurfer's [Dale et al., 1999] robust registration [Reuter et al., 2010; Wachinger and N. Navab, 2012; Wachinger et al., 2013], ANTS' SyN nonlinear registration [Avants et al., 2008], and the FSL brain extraction [Smith, 2002] tools were all incorporated in the pipeline. 3D visualization was performed in Amira (Visage Imaging, Germany).

### 2.2.1 Histology processing

#### 2.2.1.1 Blockface image segmentation

Blockface images carry considerable background clutter that must be eliminated to prevent registration errors and to reduce computational time (Figure 1 shows some examples of blockface images). Our segmentation strategy was inspired by computer vision approaches applied to human skin segmentation, where the use of a suitable orthogonal color model allows skin-colored pixels to cluster in specific regions of the color space [Yang et al., 1997]. We use the YIQ color space in our segmentation strategy, where the luminance component Y is independent of the color components I and Q. This way, lighting variations cause less interference with our color-based segmentation [Kakumanu et al., 2007]. The RGB images are converted to YIQ by the following equation:

$$\begin{bmatrix} Y \\ I \\ Q \end{bmatrix} = \begin{bmatrix} 0.299 & 0.587 & 0.114 \\ 0.596 & -0.274 & -0.322 \\ 0.211 & -0.523 & 0.312 \end{bmatrix} \begin{bmatrix} R \\ G \\ B \end{bmatrix} \quad (1)$$

Similarly to skin segmentation, color brain pixels cluster in the IQ plane [Kakumanu et al., 2007; Yang et al., 1997].

We use clustering to perform the segmentation, since it is unsupervised, eliminating the need for training on a manually annotated version of the data. A Gaussian mixture model (GMM) with two Gaussians models tissue and background distributions [Bishop, 2007]. Expectation maximization (EM) is used for estimating the GMM parameters [Duda et al., 2001]. After parameter estimation, all pixels are classified, yielding an initial binary mask. In our approach, the EM algorithm is initialized with information from pixels randomly distributed into two classes. However, users can also manually initialize it with pre-classified pixels, if preferred. The classification step can be repeated several times for segmentation improvement. In some cases the segmentation will have irregular boundaries and holes, which are corrected using an optional final refinement with active contours [Chan and L. A. Vese, 2001]. All other processing steps that use the blockface images are performed on the luminance (Y) channel, which carries a grayscale version of the image. Figure 3a shows blockface images (first row) and their segmentations using the described method.

#### 2.2.1.2 Histology segmentation

The histology datasets encompass pictures of a large, colored tissue region over a clear background, making them less complex to segment into foreground and background regions compared to the blockface images. Nevertheless, eventual unwanted background noise, such as identification marks or stickers, dust and bubbles, must be eliminated to avoid registration errors. Similar to the blockface images, all pixels belonging to the tissue of interest form clusters in the IQ plane. Therefore, we use the same strategy described above in this step. All other processing steps that use the histological images are performed on the luminance channel. Figure 3b shows the original stained histology slides before (first row) and after (second row) the segmentation.

**2.2.1.3 Histology to blockface image registration:**

We register the histological images to their blockface counterparts using an approach tailored to handle the multi-modality registration problem by running a symmetrical 2D affine registration based on robust statistics [Reuter et al., 2010]. This approach uses entropy maps created by a sliding window instead of mutual information [Wachinger et al., 2013].. Although affine registration corrects most global misalignments in our images, non-linear deformations caused by the staining process may remain. Thus, we employ a second, optional, 2D non-linear registration step using the SyN algorithm [Avants et al., 2008] to perform symmetric registration between our globally aligned inputs. We use Mattes mutual information [Mattes et al., 2003] to measure similarity. All 2D registration maps, affine and non-linear, computed in this step are stored for future use. Figure 3c shows the segmented histology slides (displayed in pseudo-color for better visualization) overlaid on their corresponding segmented blockfaces before (first row) and after (second row) registration.

**2.2.1.4 Construction of Histology 3D volume:**

Stacking the registered histological slices creates a 3D histological volume. If the registration with blockface images were successful, no z-effect remains. The voxel dimensions are calculated using a reference scale grading, which in our case was a ruler graded in millimeters. The first image in the second row of Figure 1 shows how we placed it during imaging. We performed a coarse 6 degrees of freedom (DOF) alignment of the histology volume to the *in situ* MRI. This is the only step in our pipeline that was manually executed and it will be automated in the future.

**2.2.1.5 Intensity correction:**

Intensity inconsistency among histological slices is a common occurrence caused by differences in lighting conditions and staining results. Such inconsistency causes a "stripe" effect that can be observed orthogonally to the histology sectioning plane (as shown in Figure 3d). It interferes with the visualization and analysis of anatomical structures and may also result in registration errors. We employed the iterative affine histogram-matching scheme presented by Malandain and Bardinet [Malandain and E. Bardinet, 2003] for intensity correction. Figure 3d displays a sagittal plane of the stacked histology volume before (first row) and after (second row) such correction.

### 2.2.2 MRI processing

#### 2.2.2.1 Field inhomogeneity correction

Since field intensity inhomogeneity may reduce the overall processing quality, we begin our processing with a correction step for this artifact, using the previously validated N3 algorithm [Sled et al., 1998]. Figure 3e demonstrates the intensity inhomogeneity correction process. The first row shows slices of the original MRIs, the second the inhomogeneity fields computed by the N3 algorithm and the third row the same slices after correction.

#### 2.2.2.2 Brain segmentation

This step aims to eliminate all background noise and unnecessary information (such as bones and meninges) from the images that could cause registration errors and slow down the computational processing. We perform the segmentation by using the FSL BET tool [Smith, 2002].

#### 2.2.2.3 Spatial Resampling

Histological images have higher resolution than their MRI counterparts. While digitized histology voxels have sub-millimetrical volume, a standard clinical MRI acquisition has a voxel size of around 1x1x1 mm. Direct registration of these images will cause the histology to be resampled to a lower resolution space, introducing loss of contrast and anatomical information of interest. We instead upsample the MRI volume to a higher resolution space (0.33x0.330.x0.33 mm voxel size) by employing the cubic splines interpolation. We found this resolution to be computationally tractable, while still allowing visualization of fine details in histological images, such as deep brain structures – i. e. hippocampi, substantia nigra, and caudate nucleus. Figure 3f shows the same slices before (first row) and after (second row) skull stripping.

### 2.2.3 Histology to MRI 3D registration

Three-dimensional histology to MRI registration is performed using the 3D SyN algorithm [Avants et al., 2008]. SyN is based on the large deformation diffeomorphic deformation model metric mapping (LDDMM) method [Beg et al., 2005] whose mathematical properties are especially appealing for our problem. Diffeomorphic maps are smooth and invertible functions guaranteeing that no folds or vanishing tissue will occur during the registration process. These models also enforce a 1-to-1 mapping between the movable and reference voxels and are suitable to handle large deformations [Avants et al., 2005]. SyN is a mature method that has been reported to outperform several popular registration methods [Klein et al., 2009]. Here, SyN is used with Mattes mutual information [Mattes et al., 2003] for measuring similarity, given that it is the most robust option for handling multi-modality registration problems. All 3D registration maps computed in this step are stored for future use. Figure 3h is an example of 3D histology to MRI registration using the described method. The first row shows sagittal, axial, and coronal checkerboard representations of a histology slice, overlaid on its respective MRI. The right-hand side shows a 3D rendering of one registered volume overlaid on its MRI.

### 2.2.4 Histology 3D color volume construction

Histological staining plays an important role in highlighting specific brain cytoarchitectural features. Color volume construction is an optional step in our pipeline, aiming to improve visual inspection of the registered images.

In this step, each one of the already segmented histological images (color version) is split in red, green, and blue channels, yielding three new image sets. Our pipeline reprocesses each one of these sets, starting from the "Histology to Blockface Image Registration" step (Section 2.2.1.3). However, it is not necessary to rerun the registration algorithms. Affine transforms and maps created during the 2D and 3D non-linear registration steps are reapplied to the R, G, and B channels. The three channels are recombined after the 3D registration, yielding a color volume. Figure 3i shows examples of 3D color reconstructions. Figure 4 shows a detailed diagram of the color volume construction method.

## 2.3 Quantitative evaluation

We used the *ex vivo in situ* MRI datasets (i.e. the 1 mm isotropic 3D T1-weighted acquisition performed on the deceased patient, with brain still inside the skull) as the "gold standard" to evaluate the registrations, as they did not carry deformations caused by the histological procedures.

One experienced neurosurgeon and one experienced neuroradiologist (Observer 01 and Observer 02, respectively) manually segmented the right hippocampus (RH), left caudate nucleus (LC), and lateral ventricles (VE) on the MRI datasets and intermediate histology data sets (i.e. only 2D registered to the blockface), in a slice-wise manner, on Case 01 and Case 02. All segmentations were thresholded into binary images and stacked to build 3D volumes. The 3D deformation maps, computed during the pipeline execution, were applied to these binary volumes, resulting in masks registered to their MRI volumes. We selected the above mentioned subcortical structures because they all have well-defined boundaries on MRI and histology, thus reducing segmentation time and inter-observer variability. Figure 5 shows 3D reconstructions of the stacked binary masks.

The Dice Similarity Coefficient (DSC) [Dice, 1945] was used to quantify the amount of overlap between MRI and registered histology (DSC range is in the [0, 1] interval where 0 means no overlap and 1 means maximum overlap). We additionally employed the normalized weighted spectral Distance (nWSD) [Konukoglu et al., 2012] to quantify shape dissimilarities between the MRI and histology segmentations (nWSD range is in the [0, 1] interval, where 0 means a perfect match between shapes while 1 means complete disagreement).

Both metrics were computed on the three-dimensional binary volumes created from the MRI segmentations and histology segmentation (after registration to the MRI space using the pre-computed transformation functions). This procedure was repeated for each observer.

The Jacobian determinant was used to investigate the properties of the 3D registration morphs computed during the final registration step. Positive values indicated orientation preservation, while negative values indicated orientation inversion. The mapping magnitude was given by the absolute value of the determinants [Ashburner and G. R. Ridgway, 2015].

Finally, inter-observer variability was measured by computing the DSC on the segmentations performed by Observer 01 and Observer 02.

## 3. Results

Visual inspection of the registered volumes revealed good registration quality. Figures 6 and 7 show checkerboard views of Case 01 and Case 02 histologies overlaid on their respective MRIs. They reveal overall good alignment of brain surfaces in both cases, and also matching subcortical structures in all planes. Cortical surface and white matter borders were well aligned in both cases. Qualitatively, Case 01 presented a better registration of the inner brain structures than Case 02. This was especially visible on the lateral ventricles. Inspection of the 3D reconstruction also revealed good overall registration quality. Figure 8 shows the final 3D reconstruction of Case 01 and Case 02. Consistent inner brain structure shape was visible in both reconstructions.

Quantitative evaluation using DSC and nWSD also revealed good localization and shape agreement between registered images and their corresponding ground truth MRI. Table I summarizes the results obtained by computing DSC and nWSD.

Most DSC values were above 0.7 for Case 01 and Case 02, indicating good localization of the segmented structures [Crum et al., 2006]. Exceptions were Case 02 LC (by Observer 01), RH (by Observer 02), and Case 01 RH (by Observer 01) whose DSCs were below 0.7. Table I also shows that all structures had low nWSD (below 0.01) indicating high overall 3D shape agreement between the gold standard and registered histology. In summary, all segmented structures were well aligned with the ground truth and also presented good shape agreement among them.

We also computed the inter-observer DSC to verify the amount of agreement between the gold standard segmentation sets. Column "MRI" in Table II shows the amount of overlap between segmentations performed by Observer 01 and Observer 02 on the MRI set, for each considered structure. Column "histology" shows the same information, but for segmentations on the histology set. All DSC coefficients are higher than 0.75, indicating a high level of agreement between observers [Crum et al., 2006].

Finally, the Jacobian determinant analysis of the 3D registration fields yielded values ranging from 0 to 3.4 for Case 01 and 0 to 3.18 for Case 02. Figure 10 displays the axial, coronal, and sagittal views of the Jacobian determinants. Darker regions show where the lower determinant values are located, indicating that these regions underwent smaller displacements.

## 4. Discussion

We developed an image processing pipeline to compute spatial correspondence between whole human brain histology and correposnding MRI images. This pipeline was designed to streamline the registration of histology images to MRI, relying on minimal user interaction. Our proposed pipeline achieved very good precision according to visual inspection and quantitative coefficients. Although it was tailored to automate our histology to MRI previously-published registration protocol [Grinberg et al., 2008], it can be easily adapted to other registration protocols, since the selected algorithms are suitable to run on conventional machines (instead of High-Performance Computer systems) and it relies on a standard clinical MRI acquisition. In this paper we described all steps in depth to facilitate other groups to implement this pipeline.

To quantify the correspondence between the histological and MRI volumes, we calculated the amount of overlap between regions of interest traced by experts separately onto the MRI and histology. DSC is a non-parametric statistical metric, which measures the amount of overlap between two segmented areas. It has an intuitive meaning and it is computationally inexpensive, thus being widely used in segmentation and registration studies. Our average DSC value was higher than 0.7 (Table I). Previous registration studies suggest that DSC values above 0.60 and 0.70 are highly satisfactory for small and large brain structures, respectively [Akbarzadeh et al., 2013; Dawant et al., 1999; Pasquier et al., 2007; Sparks et al., 2002]. We obtained an average DSC of 0.69 for the right hippocampus - a small structure - and 0.74 for the left caudate - a large structure. These results suggest that our pipeline is satisfactory regarding spatial co-localization.

As DSC measures overlap between two sets, rather than information about shape differences, we also used a second metric, the nWSD, to measure the overall geometric similarity between the MRI and histological volumes. Authors have shown that nWSD values vary approximately in line with shape dissimilarity and that values above 0.01 reflect significant differences [Konukoglu et al., 2012]. In our study, nWSD values computed for RH, LC, and VE were below 0.001 (Table I) for both brains, suggesting that our pipeline is suitable for recovering most of the brain's original geometry in deformed histological datasets. Although nWSD is a relatively new metric and is computationally expensive, we believe that analyzing DSC and WSD values side-by-side provides complementary information to evaluate the usefulness of the pipeline.

To address if inter-observer variability in segmenting the regions of interest could have impacted DSC values or quality assessment of our study, we calculated the DSC beween observers in histology and MRI images for each case (Table II). We obtained DSCs varying from 0.75 to 0.91 indicating a high level of agreement between observers [Crum et al., 2006].

We used the Jacobian determinant computed over the deformation maps to investigate our registration properties. Positive determinant values indicate tissue expansion, while negative values indicate retraction and folding. Zero values indicate vanishing tissue. We did not obtain negative values, indicating the absence of folding during the registration process, which is in accordance with diffeomorphic mapping. Although many registration models allow the occurrence of folding [Ou et al., 2012], it is an unrealistic deformation for a human organ (even for a fixed brain) [Rey et al., 2002], therefore not desirable. It is interesting to note the existence of zero values in some regions. In theory, it would mean tissue disappearance. However, SyN uses finite differences to compute the determinants, a method known to present numerical problems with diffeomorphic mappings that lead to inaccurate results. Thus, zero values found here are more likely to indicate contractions that were beyond the capture range of finite differences than actual tissue information loss [Dupuis et al., 1998].

Any co-registration pipeline can be impacted by the quality of the histological procedure and tissue quality, as illustrated here by comparing Case 01 and Case 02 using visual inspection and quantitative coefficients (Figures 6-9).

First, Case 02 was sliced in the sagittal plane whereas Case 01 was sliced in the axial plane (see Figure 1). Sagittal blockface images resulted in more "transparency artifacts" caused by a large empty space filled by translucid mounting media inside the borders. We call "transparency artifacts" regions of the brain that do not belong to a particular slice but appear on its blockface image due to the mounting media's transparency (i.e. tissue that is located below the actual slice plane but that is still visible). Transparency artifacts were present in both cases but were pervasive in Case 02, especially around the lateral sulcus. In many cases these artifacts are barely distinguishable from tissue actually located in the current slide (even by human observers), as illustrated in Figure 1e. Artifacts cause registration errors, such as excessive scaling or unnecessary rotations, thus yielding poor quality histology volumes. We addressed this problem by dyeing the mounting media blue to minimize artifacts [Theofilas et al., 2014]. The overall tissue quality also jeopardizes the registration. Histological slides from Case 02 bore small holes and tears, along with a huge tear in the frontal cortex (visible in Figure 1d, as a large gap on the frontal cortex and Figure 2m) produced during the fixation process. These artifacts are unlikely to be properly corrected by the 2D and 3D registration pipelines [Annese et al., 2014; Lepage et al., 2010; Lewis et al., 2014]. Optimizing the histological procedure is key for obtaining better registration. We and others conducted extensive experiments to optimize histological tissue handling for histology to MRI registration studies [Augustinack et al., 2010; Grinberg et al., 2008]. Nevertheless, the quality of the tissue is often unpredictable and depends on preagonal conditions and postmortem interval. When possible, these parameters need to be considered for case selection.

Future studies will focus in overcoming the remaining limitations of our pipeline. First, we do not have an automated method for correcting some of the histology artifacts noted. Similar to other authors [Amunts et al., 2013], we have performed manual correction to address tissue dislocation (fractures during the histological processing). At this point, the pipeline uses a method limited to correct small holes [Lepage et al., 2010]. We are also investigating if the use of whole brain slides, as opposed to hemispheric slides as proposed initially, reduces the deformations sufficiently to allow a direct registration between the stained histological slides to MRI, skipping the blockface as an intermediate registration step. It is yet to be seen if this pipeline provides precise registration for other brain structures. Finally, we are preparing whole brain histological datasets using blue mounting media, aiming to minimize artifacts that jeopardize the registration.

Our pipeline allowed the direct comparison of signals obtained in MRI to their microscopic features in the brain tissue. Although designed for whole brain images, our method does not rely on any particular anatomical feature. It is flexible enough to be used in small tissue samples as long as the image modalities are similar to those used here. Other laboratories may be able to benefit from our method to build platforms for correlating clinical, imaging, and histological (and immune-histo-chemical) data.

**Acknowledgements:** This study was funded by CNPq (grant 400698/2012-9), IIEP Albert Einstein, Federico Foundation, and NIH 1R01NS070963-01A1, R01AG040311, LTG is also supported by K24AG053435, P50AG023501, P01AG019724.

|      |           | Observer 01 |        | Observer 02 |        | Mean        |               |
| ---- | --------- | ----------- | ------ | ----------- | ------ | ----------- | ------------- |
| Case | Structure | DSC         | nWSD   | DSC         | nWSD   | DSC         | nWSD          |
| 01   | RH        | 0.68        | 0.0017 | 0.70        | 0.0050 | 0.69        | 0.0034        |
|      | LC        | 0.80        | 0.0020 | 0.78        | 0.0020 | 0.79        | 0.0020        |
|      | VE        | 0.76        | 0.0014 | 0.78        | 0.0015 | 0.77        | 0.0014        |
| mean |           | **0.75**    | **0.0017** | **0.75** | **0.0028** | **0.75** | **0.0023** |
| 02   | RH        | 0.72        | 0.0007 | 0.67        | 0.0017 | 0.69        | 0.0012        |
|      | LC        | 0.65        | 0.0044 | 0.76        | 0.0070 | 0.70        | 0.0057        |
|      | VE        | 0.75        | 0.0022 | 0.75        | 0.0020 | 0.75        | 0.0021        |
| mean |           | **0.71**    | **0.0024** | **0.73** | **0.0036** | **0.71** | **0.0030** |
| total mean |     |             |        |             |        | 0.73±0.0440 | 0.0026±0.0017 |

**Table** I. 3D Dice similarity coefficient (DSC) and 3D normalized Weighted Similarity Distance (nWSD) values computed using manual segmentation of the right hippocampus (RH), left caudate nucleus (LC), and lateral ventricles (VE).

| Case | Structure | MRI  | Histology |
| ---- | --------- | ---- | --------- |
| 01   | RH        | 0.82 | 0.86      |
|      | LC        | 0.76 | 0.9       |
|      | VE        | 0.9  | 0.91      |
| 02   | RH        | 0.81 | 0.85      |
|      | LC        | 0.81 | 0.83      |
|      | VE        | 0.90 | 0.86      |

**Table II:** DSC values computed between the binary masks created from the manual segmentations performed by two independent observers.

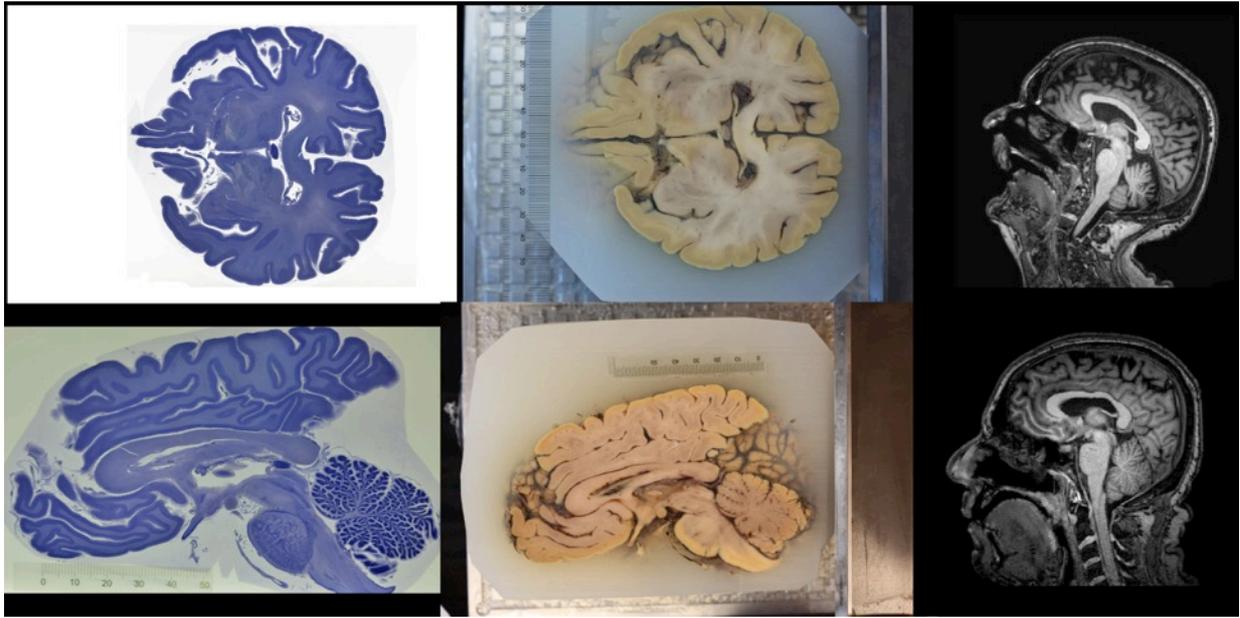

**Figure 1.** Example of original images from Case 01 (first line) and Case 02 (second line). First column: gallocyanin stained histology. Second column: their corresponding blockface images. Third column: MRI slices. 1e is a good example of transparency artifact: in that slice frontal and occipital cortices are visible even though they do not belong to that particular slice.

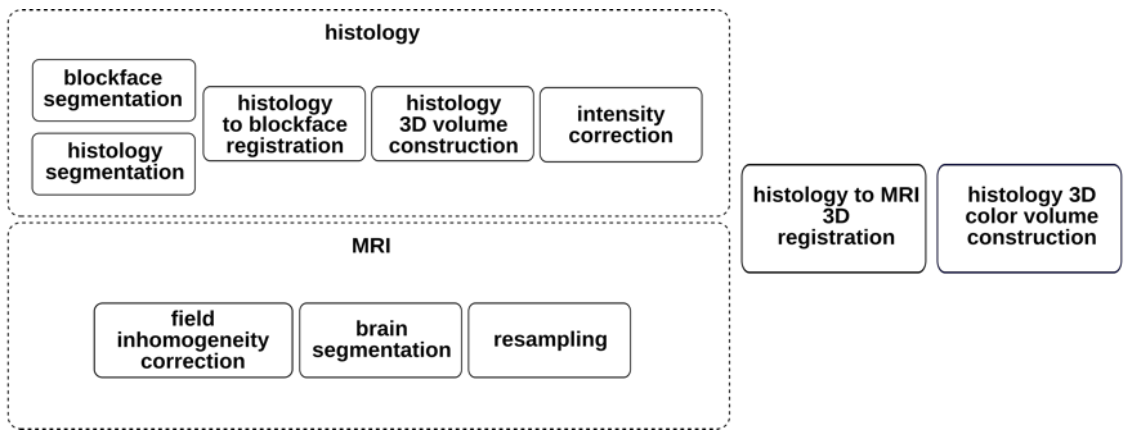

**Figure 2.** Our proposed computational pipeline.

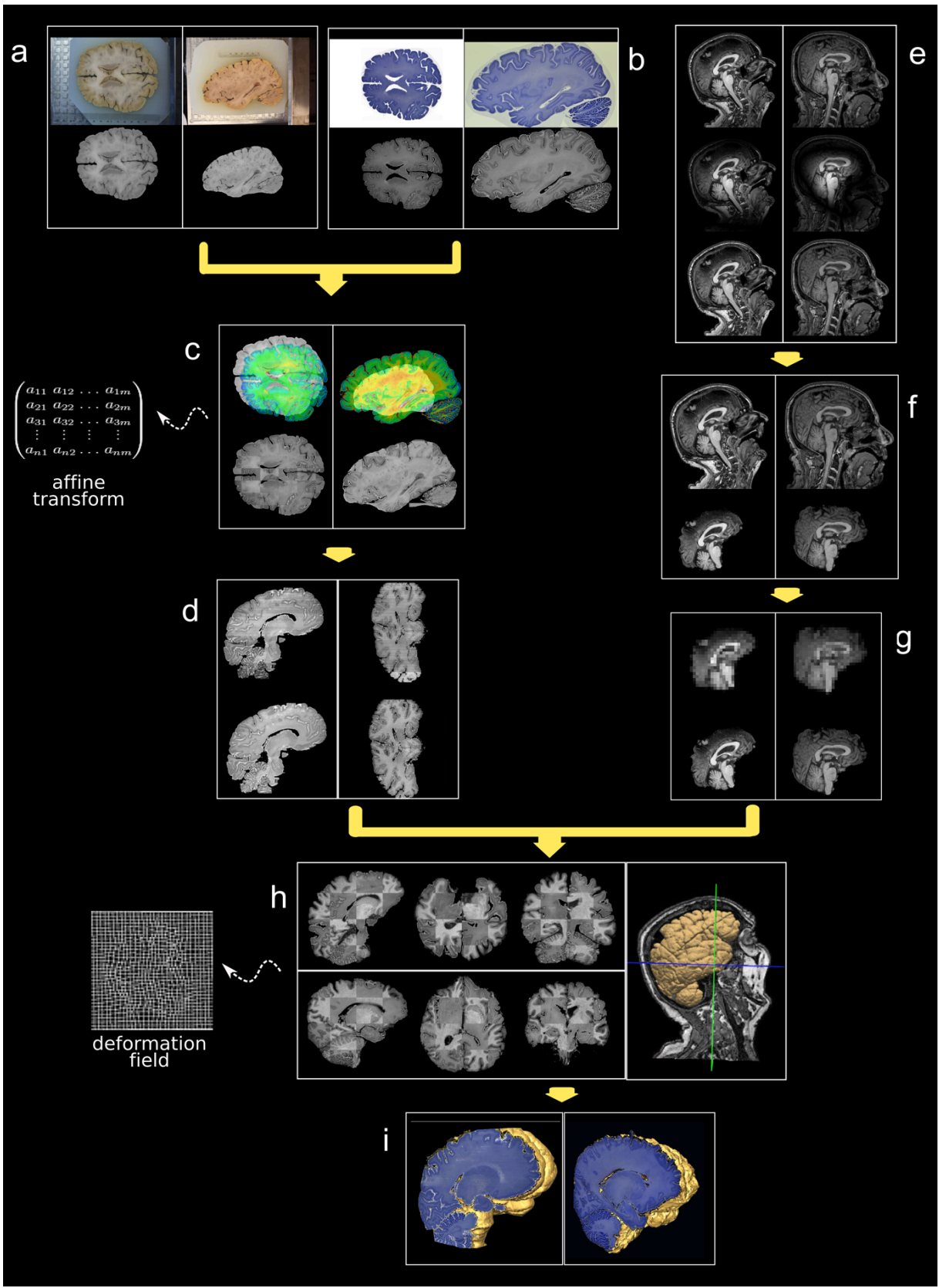

**Figure 3**. Step-by-step pipeline processing results. In 3a-3g: Case 01 images are always displayed in the left column while Case 02 images are always displayed in the right column. a) Blockface images (first row) and their corresponding segmentations after processing; b) same slides after staining with gallocyanin (first row) and after segmentation; c) shows the same histology slides (displayed in pseudo-color for better visualization) overlaid on their corresponding blockfaces (first row) and after 2D registration. Scale and shape differences between histology and blockface are evident in these examples and are corrected by the 2D registration; d) sagittal cut of the stacked histology volume before intensity correction (first row) and after intensity correction. Differences on the slice intensities appear as stripes with different intensities, scattered around the volume on the first row; e) MRI intensity inhomogeneity correction: first row shows slices of the original MRI, second row shows the inhomogeneity fields computed by the N3 algorithm, third row shows the same slices after correction; f) same MRI slices before (first row) and after skull stripping; g) example upsampling process; h) histology registered to MRI. First row shows sagittal, axial, and coronal checkerboard representations of a Case 01 histology slice, overlaid on its respective MRI. Second row shows same views of Case 02. The right shows the 3D rendering of the Case 01 registered histology over its non-segmented MRI.; i) Case 01 (left) and Case 02 (right) final 3D color reconstructions.

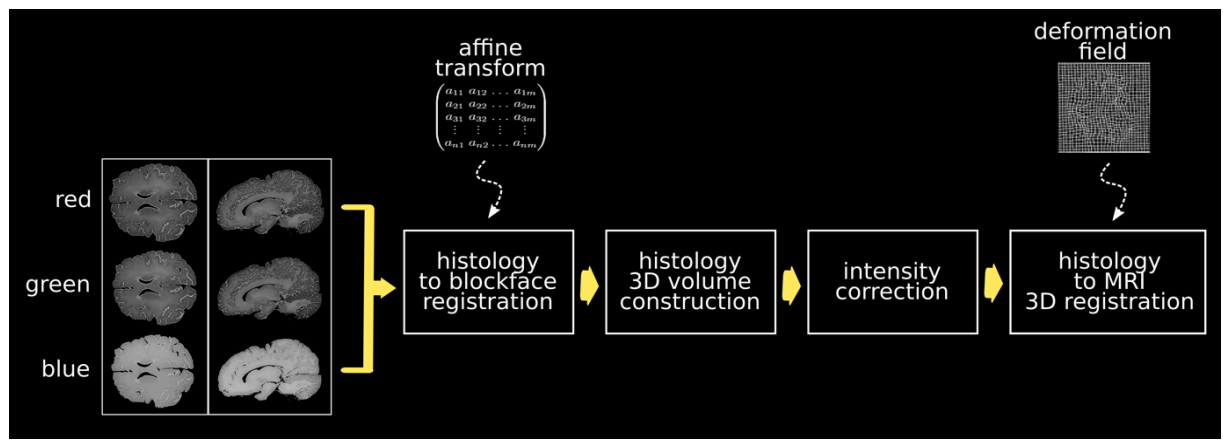

**Figure 4**: Detailed view of the 3D color reconstruction step.

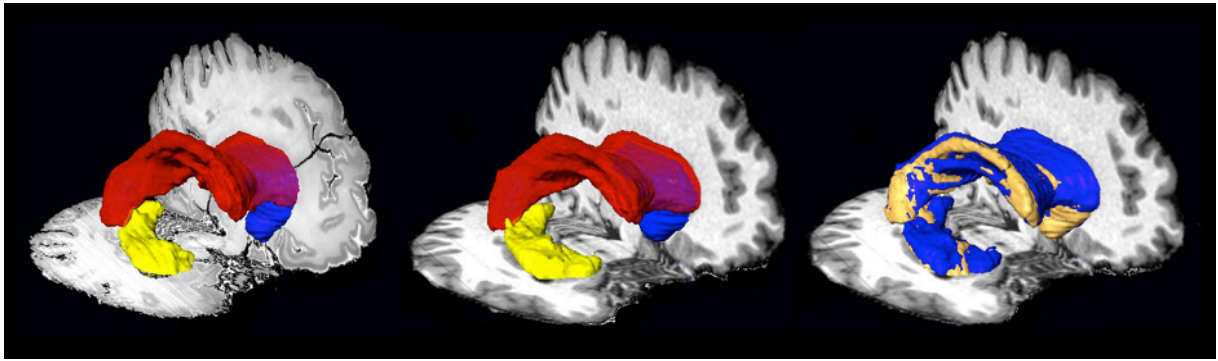

**Figure 5.** 3D reconstruction of Case 02 binary masks used for quantifying registration quality. Left: histology segmentation of the ventricles (red), right hippocampus (yellow), and left caudate nucleus (blue). Middle: the same, for the MRI volume. Right: segmented histology structures (yellow) overlaid by the segmented MRI structures (blue).

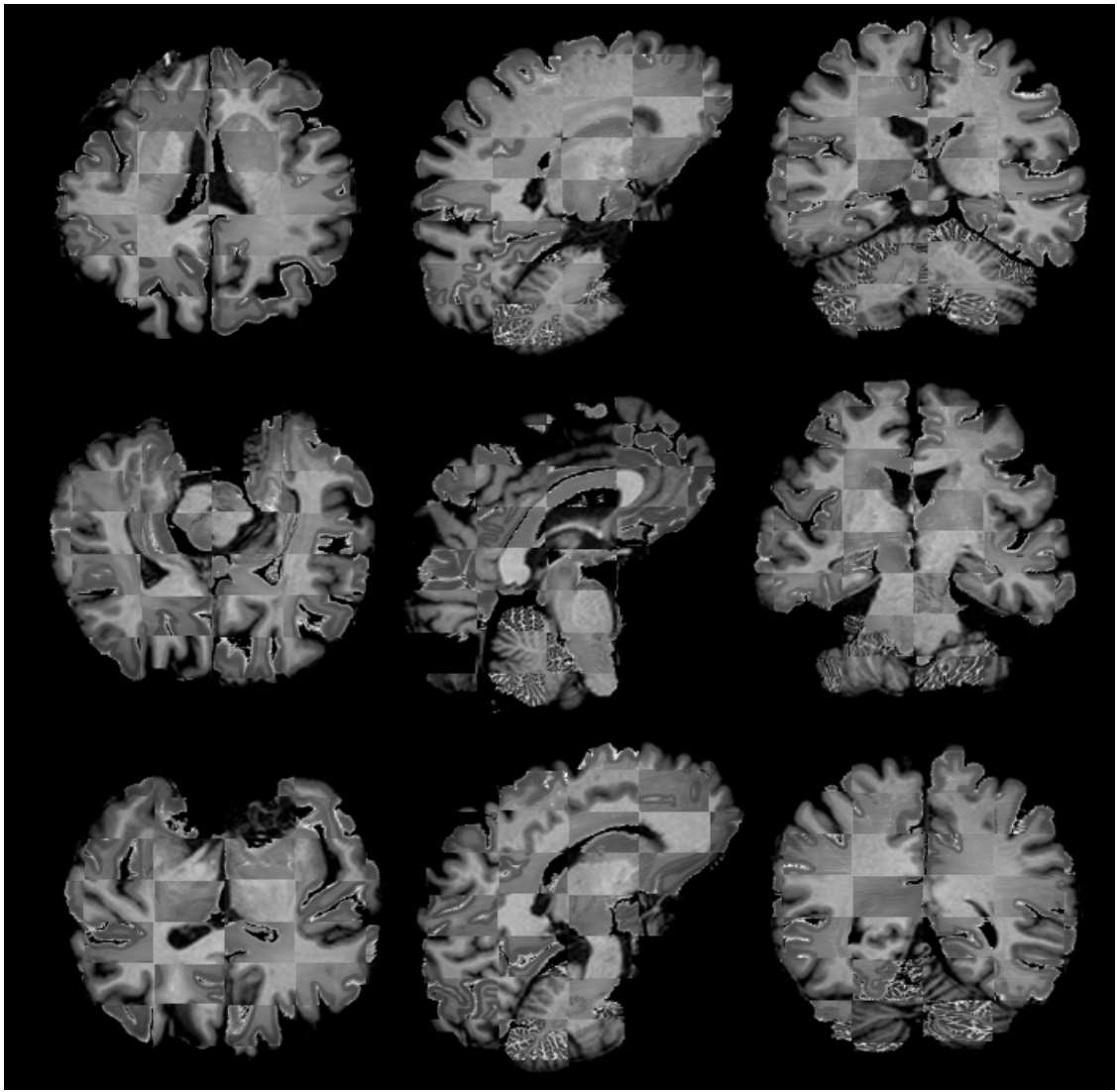

**Figure 6:** Case 01 registration results. Each row shows the axial, sagittal, and coronal view of a different slice in Case 01. The histology was overlaid on the MRI in a "checkerboard" pattern for easier visualization.

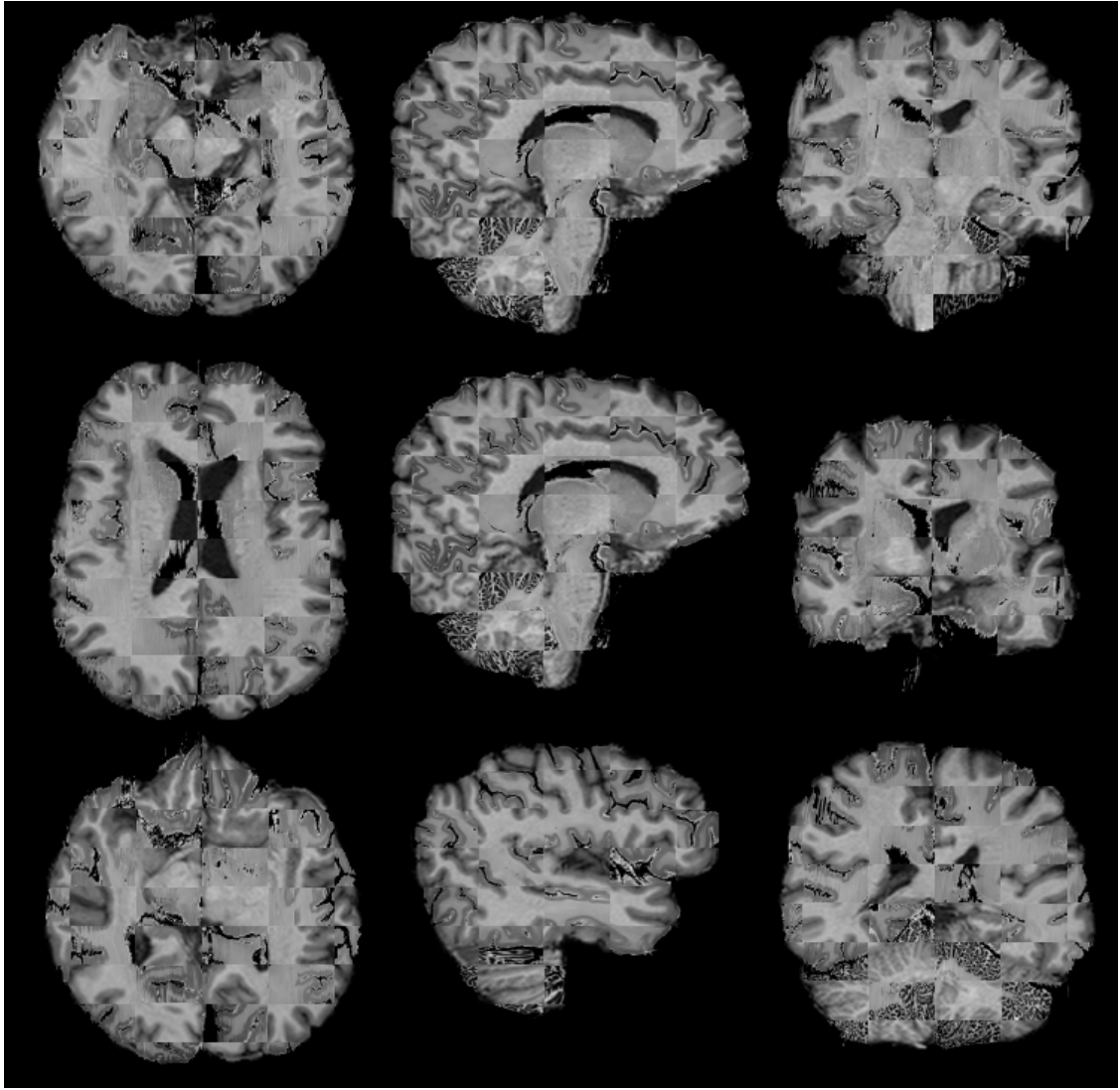

**Figure 7**: Case 02 registration results. Each row shows the axial, sagittal, and coronal view of a different slice in Case 02. The histology was overlaid on the MRI in a "checkerboard" pattern for easier visualization.

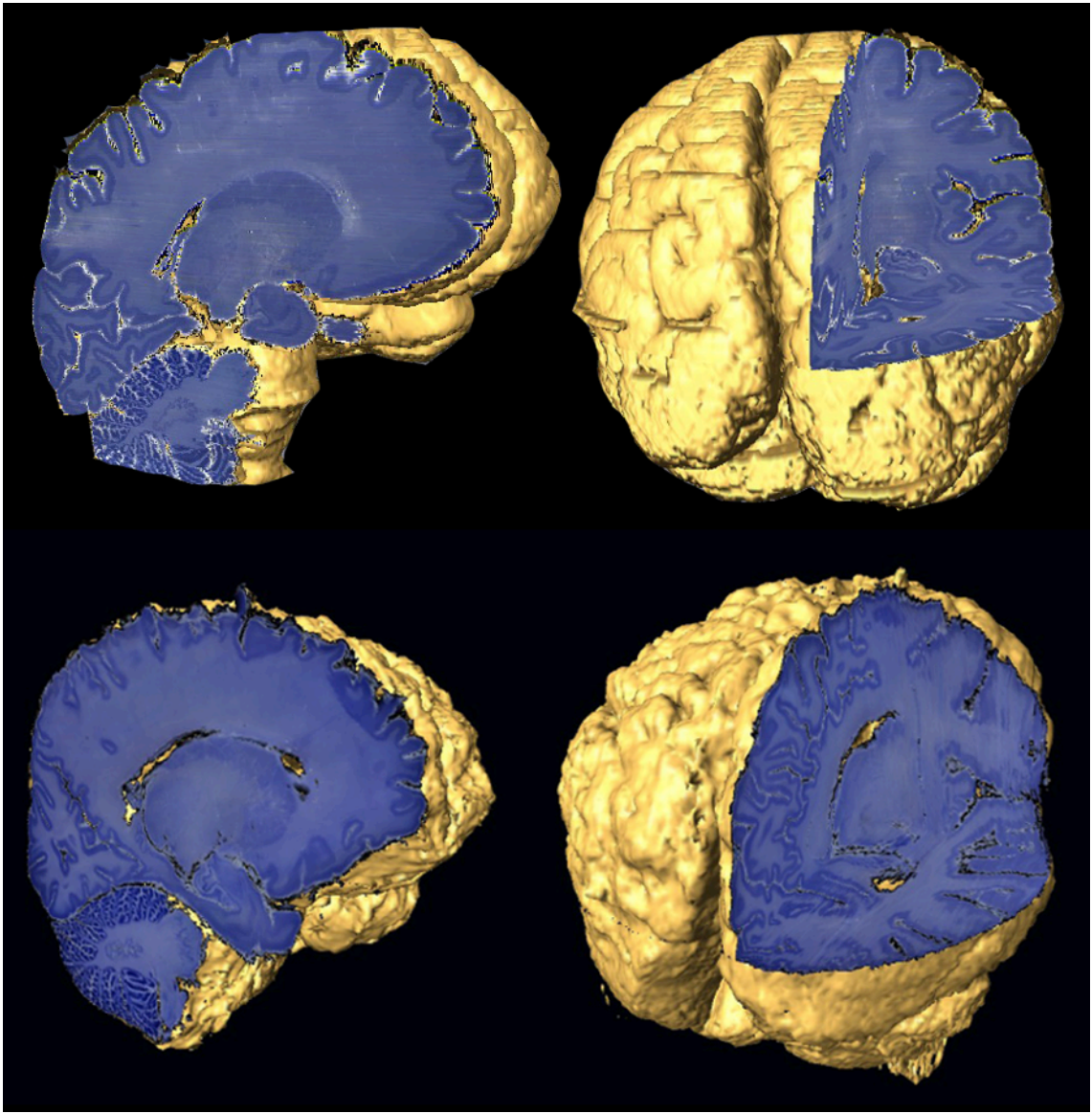

**Figure 8**. Final 3D color reconstructions. Top row: two views of Case 01. Bottom row: two views of Case 02.

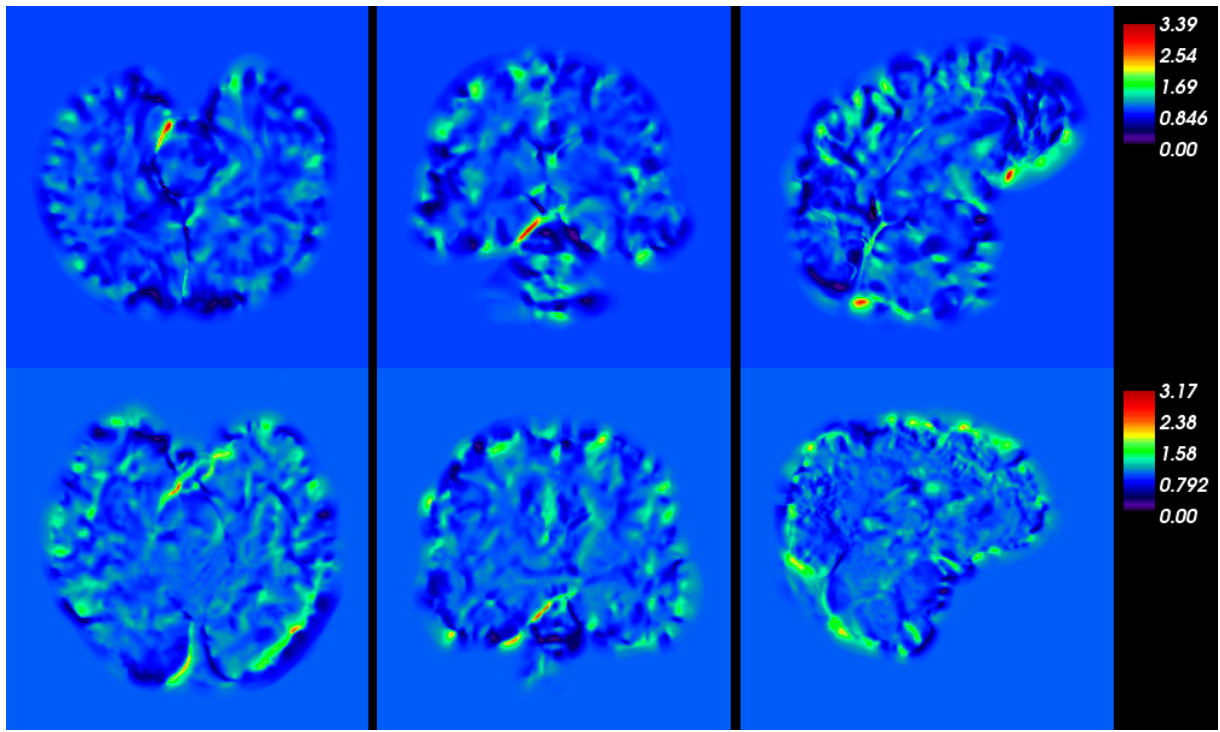

**Figure 9**: First row: Axial, coronal, and sagittal views of the Jacobian determinant of the 3D registration transform used to warp Case 01. Second row: Axial, coronal, and sagittal views of the Jacobian determinant of the 3D registration transform used to warp Case 02. Regions with brighter colors (green and red) had greater displacements applied to them.